%% file: neurips_2019.tex
\documentclass{article}

     \usepackage[final]{neurips_2019}

\usepackage[utf8]{inputenc} 
\usepackage[T1]{fontenc}    
\usepackage{hyperref}       
\usepackage{url}            
\usepackage{booktabs}       
\usepackage{amsfonts}       
\usepackage{nicefrac}       
\usepackage{microtype}      

\usepackage{graphicx}
\usepackage{subfigure}
\usepackage{booktabs} 
\usepackage{tikz}
\usepackage{pgfpages}
\usepackage{pgfplots}

\usepackage{algorithm}
\usepackage{algorithmic}
\usepackage{graphicx}
\usepackage{textcomp}
\usepackage{xcolor}
\usepackage{amsmath}
\usepackage{lipsum}
\usepackage{mathtools}
\usepackage{float}
\usepackage{cuted}
\usepackage{amssymb}

\usepackage{amsmath,amssymb,amsfonts}
\usepackage{color}

\usepackage{textcomp}
\usepackage{xcolor}
\usepackage{amsmath}
\usepackage{lipsum}
\usepackage{mathtools}
\usepackage{float}
\usepackage{cuted}

\title{Attention Based Pruning for Shift Networks}

\author{Ghouthi Boukli Hacene$^{1,2}$, Carlos Lassance$^{1,2}$, Vincent Gripon $^{1,2}$\\ \textbf{Matthieu Courbariaux}$^1$ and \textbf{Yoshua Bengio}$^1$ \\
 $^1$ Universit\'e de Montr\'eal, MILA, $^2$IMT Atlantique, Lab-STICC\\
}

\begin{document}

\maketitle

\begin{abstract}
In many application domains such as computer vision, Convolutional Layers (CLs) are key to the accuracy of deep learning methods. However, it is often required to assemble a large number of CLs, each containing thousands of parameters, in order to reach state-of-the-art accuracy, thus resulting in complex and demanding systems that are poorly fitted to resource-limited devices. Recently, methods have been proposed to replace the generic convolution operator by the combination of a shift operation and a simpler 1x1 convolution. The resulting block, called Shift Layer (SL), is an efficient alternative to CLs in the sense it allows to reach similar accuracies on various tasks with faster computations and fewer parameters. In this contribution, we introduce Shift Attention Layers (SALs), which extend SLs by using an attention mechanism that learns which shifts are the best at the same time the network function is trained. We demonstrate SALs are able to outperform vanilla SLs (and CLs) on various object recognition benchmarks while significantly reducing the number of float operations and parameters for the inference.
\end{abstract}

\section{Introduction}
\label{sec:introduction}
Convolutional Neural Networks (CNNs) are the state-of-the-art in many computer vision tasks, such as image classification, object detection and face recognition~\cite{szegedy2016rethinking}. To achieve top-rank accuracy, CNNs rely on the use of a large number of trainable parameters, and considerable computational complexity. This is why there has been a lot of interest in the past few years towards the compression of CNNs, so that they can be deployed onto embedded systems. The purpose of compressing DNNs is to reduce memory footprint and/or computational complexity while mitigating losses in accuracy.

Prominent works to compress neural networks include binarizing (or quantifying) weights and activations~\cite{courbariaux2015binaryconnect,soulie2016compression,rastegari2016xnor,lin2015neural,li2016ternary,zhou2016dorefa,han2015learning,wu2018deep}, pruning network connections during or before training~\cite{han2015learning,han2015deep,li2016pruning,ardakani2016sparsely}, using grouped convolutions~\cite{chollet2017xception,howard2017mobilenets,sandler2018mobilenetv2}, introducing new components~\cite{zhang2018interpretable,juefei2017local}, using convolutional decomposition~\cite{szegedy2015rethinking,simonyan2014very}, or searching for a lightweight neural network architecture~\cite{iandola2016squeezenet}.

Recently, the authors of~\cite{wu2017shift} have proposed to replace the generic convolution operator with the combination of shifts and simpler 1x1 convolutions. In their work, the shifts are hand-crafted and all the parameters to be trained are in the 1x1 convolutions. These operations are particularly well suited to embedded devices~\cite{hacene2018quantized,yang2018synetgy}.

In this paper, we introduce the Shift Attention Layer (SAL), which can be seen as a selective shift layer. Indeed, the proposed layer starts with a vanilla convolution and learns to transform it into a shift layer throughout the training of the network function. It uses an attention mechanism~\cite{vaswani2017attention} that selects the best shift for each feature map of the architecture, what can be seen as a pruning technique. We demonstrate it is able to significantly outperform original shift layers~\cite{wu2017shift} and other pruning techniques at the cost of requiring more parameters during training, but ends up with less parameters for inference. It is thus of particular interest for implementing the inference process on resource-limited systems.

The outline of the paper is as follows. In Section~\ref{sec:methodology} we explain the proposed method. In Section~\ref{sec:relatedwork} we present related work. In Section~\ref{sec:experiments} we present experiments on challenging computer vision datasets. Finally, In Section~\ref{sec:conclusion} we discuss future work and conclude. All the code used to run the experiments in this paper is available online:~\url{https://hiddenduringreviewing}.

\begin{figure}
    \centering
    \begin{tikzpicture}[scale=0.6]
        \def \r{.65}
        \draw (0,0)--(2.5,0)--(2.5,5)--(0,5)--cycle;
        \draw[<->](-.5,0)--(-.5,5);
    \node[text width=3cm] at (.8+\r,2.5) 
         {$L$};
         
    \draw[<->](0,-0.5)--(2.5,-.5);
    \node[text width=3cm] at (2.9+\r,-.9) 
         {$C$};
         \node[text width=3cm] at (2.9+\r,-2.1) 
         {$(1)$};
        \foreach \x in {.5,1,1.5,2}{\draw (\x,0)--(\x,5);}
        \foreach \y in {.5,1,1.5,2,2.5,3,3.5,4,4.5}{\draw (0,\y)--(2.5,\y);}

        \draw[very thick] (0,3.5)--(2.5,3.5)--(2.5,2)--(0,2)--cycle;
     \draw[<->](3,3.2)--(3,2.2);
    \node[text width=3cm] at (5+\r,2.8) 
         {$S$};
         
        \draw (4.5,0)--(5,0)--(5,5)--(4.5,5)--cycle;
        \foreach \y in {.5,1,1.5,2,2.5,3,3.5,4,4.5}{\draw (4.5,\y)--(5,\y);}
        \draw[very thick](4.5,3)--(5,3)--(5,2.5)--(4.5,2.5)--cycle;
        
        \draw (2.5,3.5)--(4.5,3);
        \draw (2.5,2)--(4.5,2.5);
        \draw[->](3.5,3.4)--((3.5,4);
        \draw[->](3.5,2.1)--((3.5,1.5);
        
        \draw[->](5.2,3.1)--((5.2,3.7);
        \draw[->](5.2,2.4)--((5.2,1.8);
        
        \node[text width=3cm] at (2.9+\r,5.3) 
         {$\mathbf{X}$};
         
        \node[text width=3cm] at (6.4+\r,5.3) 
         {$\mathbf{Y}_d$};
         
         
         \def \x1{8}
         
                 \draw (0+\x1,0)--(2.5+\x1,0)--(2.5+\x1,5)--(0+\x1,5)--cycle;
        \draw[<->](-.5+\x1,0)--(-.5+\x1,5);
    \node[text width=3cm] at (.8+\x1+\r,2.5) 
         {$L$};
         
    \draw[<->](0+\x1,-0.5)--(2.5+\x1,-.5);
    \node[text width=3cm] at (2.9+\x1+\r,-.9) 
         {$C$};
         \node[text width=3cm] at (2.9+\x1+\r,-2.1) 
         {$(2)$};
         
        \foreach \x in {8.5,9,9.5,10}{\draw (\x,0)--(\x,5);}
        \foreach \y in {.5,1,1.5,2,2.5,3,3.5,4,4.5}{\draw (0+\x1,\y)--(2.5+\x1,\y);}

        \draw[very thick] (0+\x1,3.5)--(2.5+\x1,3.5)--(2.5+\x1,2)--(0+\x1,2)--cycle;
     \draw[<->](3+\x1,3.2)--(3+\x1,2.2);
    \node[text width=3cm] at (5+\x1+\r,2.8) 
         {$S$};
         
        \draw (4.5+\x1,0)--(5+\x1,0)--(5+\x1,5)--(4.5+\x1,5)--cycle;
        \foreach \y in {.5,1,1.5,2,2.5,3,3.5,4,4.5}{\draw (4.5+\x1,\y)--(5+\x1,\y);}
        \draw[very thick](4.5+\x1,3)--(5+\x1,3)--(5+\x1,2.5)--(4.5+\x1,2.5)--cycle;
        
        \draw (2.5+\x1,3.5)--(4.5+\x1,3);
        \draw (2.5+\x1,2)--(4.5+\x1,2.5);
        \draw[->](3.5+\x1,3.4)--((3.5+\x1,4);
        \draw[->](3.5+\x1,2.1)--((3.5+\x1,1.5);
        
        \draw[->](5.2+\x1,3.1)--((5.2+\x1,3.7);
        \draw[->](5.2+\x1,2.4)--((5.2+\x1,1.8);
        
        \node[text width=3cm] at (2.9+\x1+\r,5.3) 
         {$\mathbf{X}$};
         
        \node[text width=3cm] at (6.4+\x1+\r,5.3) 
         {$\mathbf{Y}_d$};

         \def \y{3.5}
         \def\x{{8,8.5,9,9.5,10}}
         \def\c{{0.9, 0.15, 0.75, 0 , 0.12,
       0.22, 0.1, 0.6, 0.22, 0.44 ,
       0.18470363, 0.16826004
}}
    \foreach \i in {0,...,4}{
    
    \draw [fill opacity=\c[\i), fill=black] (0+\x[\i],0+\y)--(.5+\x[\i],0+\y)--(.5+\x[\i],-.5+\y)--(0+\x[\i],-.5+\y)--cycle;}

     \def \y{3}
         \def\x{{8,8.5,9,9.5,10}}
         \def\c{{0.05, 0.05, 0.1, 0.8 , 0.12,
       0.22, 0.1, 0.6, 0.22, 0.44 ,
       0.18470363, 0.16826004
}}
    \foreach \i in {0,...,4}{
    
    \draw [fill opacity=\c[\i), fill=black] (0+\x[\i],0+\y)--(.5+\x[\i],0+\y)--(.5+\x[\i],-.5+\y)--(0+\x[\i],-.5+\y)--cycle;}

     \def \y{2.5}
         \def\x{{8,8.5,9,9.5,10}}
         \def\c{{0.05, 0.8, 0.15, 0.2 , 0.86,
       0.22, 0.1, 0.6, 0.22, 0.44 ,
       0.18470363, 0.16826004
}}
    \foreach \i in {0,...,4}{
    
    \draw [fill opacity=\c[\i), fill=black] (0+\x[\i],0+\y)--(.5+\x[\i],0+\y)--(.5+\x[\i],-.5+\y)--(0+\x[\i],-.5+\y)--cycle;}
    
    \def \x2{16}
    \foreach \y in {-.5,0,.5,1,1.5,2,2.5,3,3.5,4,4.5}{\draw (16,\y)--(16.5,\y);}
    \foreach \y in {.5,1,1.5,2,2.5,3,3.5,4,4.5,5,5.5}{\draw (16.5,\y)--(17,\y);}
    \foreach \y in {-.5,0,.5,1,1.5,2,2.5,3,3.5,4,4.5}{\draw (17,\y)--(17.5,\y);}
    \foreach \y in {0,.5,1,1.5,2,2.5,3,3.5,4,4.5,5}{\draw (17.5,\y)--(18,\y);}
    \foreach \y in {.5,1,1.5,2,2.5,3,3.5,4,4.5,5,5.5}{\draw (18,\y)--(18.5,\y);}
    
    \draw(16,-.5)--(16,4.5);
    \draw(16.5,-.5)--(16.5,5.5);
    \draw(17,-.5)--(17,5.5);
    \draw(17.5,-.5)--(17.5,5);
    \draw(18,0)--(18,5.5);
    \draw(18.5,0.5)--(18.5,5.5);
    \draw[very thick] (16,3)--(16+2.5,3)--(16+2.5,2.5)--(16,2.5)--cycle;
    \draw[<->](-.5+16,-.5)--(-.5+16,4.5);
    \node[text width=3cm] at (.8+16+\r,2.5) 
         {$L$};
         \draw (4.5+\x2,0)--(5+\x2,0)--(5+\x2,5)--(4.5+\x2,5)--cycle;
        \foreach \y in {.5,1,1.5,2,2.5,3,3.5,4,4.5}{\draw (4.5+\x2,\y)--(5+\x2,\y);}
        \draw[very thick](4.5+\x2,3)--(5+\x2,3)--(5+\x2,2.5)--(4.5+\x2,2.5)--cycle;
        
        \draw (16+2.5,3)--(4.5+\x2,3);
        \draw (16+2.5,2.5)--(4.5+\x2,2.5);
       
        \draw[->](3.5+\x2,3.1)--((3.5+\x2,3.7);
        \draw[->](3.5+\x2,2.4)--((3.5+\x2,1.8);
        
        \draw[->](5.2+\x2,3.1)--((5.2+\x2,3.7);
        \draw[->](5.2+\x2,2.4)--((5.2+\x2,1.8);
        
        \node[text width=3cm] at (2.6+\x2,5.9) 
         {Shifted $\mathbf{X}$};
         
        \node[text width=3cm] at (6.4+\x2+\r,5.3) 
         {$\mathbf{Y}_d$};
         \draw[<->](0+\x2,-1)--(2.5+\x2,-1);
    \node[text width=3cm] at (2.9+\x2+\r,-1.4) 
         {$C$};
         \node[text width=3cm] at (2.9+\x2+\r,-2.1) 
         {$(3)$};
         
    \end{tikzpicture}
    \vspace{-.5cm}
    \caption{Overview of the proposed method: we depict here the computation for a single output feature map $d$, considering a 1d convolution and its associated shift version. Panel (1) represents a standard convolutional operation: the weight filter $\mathbf{W}_{d,\cdot,\cdot}$ containing $S C$ weights is moved along the spatial dimension ($L$) of the input to produce each output in $\mathbf{Y}_d$. In panel (2), we depict the attention tensor $\mathbf{A}$ on top of the weight filter: the darker the cell, the most important the corresponding weight has been identified to be. At the end of the training process, $\mathbf{A}$ should contain only binary values with a single 1 per slice $\mathbf{A}_{d,c,\cdot}$. In panel (3), we depict the corresponding obtained shift layer: for each slice along the input feature maps ($C$), the cell with the highest attention is kept and the others are disregarded. As a consequence, the initial convolution with a kernel size $S$ has been replaced by a convolution with a kernel size 1 on a shifted version of the input $\mathbf{X}$. As such, the resulting operation in panel (3) is exactly the same as the shift layer introduced in~\cite{wu2017shift}, but here the shifts have been trained instead of being arbitrarily predetermined.}
    \label{fig:convolution}
    \vspace{-.2cm}
\end{figure}
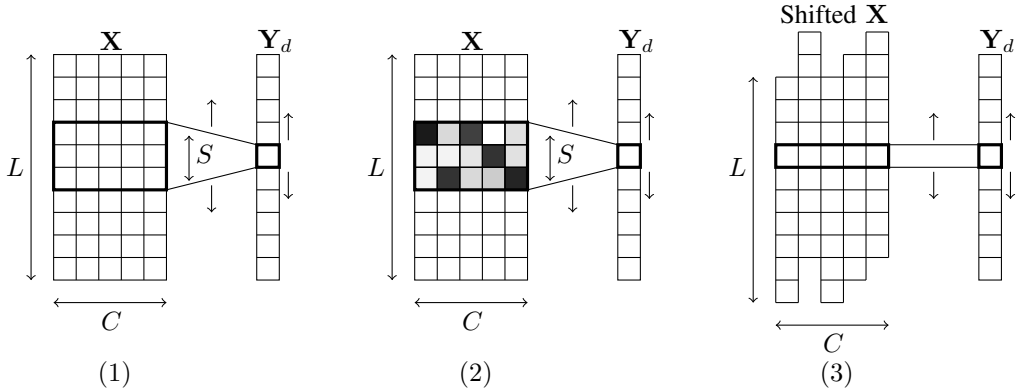

\section{Methodology}
\label{sec:methodology}
In this section, we first review the classical spatial convolution operation and see how it can be related to the shift operation defined in~\cite{wu2017shift}. We then introduce our proposed method.

\subsection{Convolution/Shift operation}

Let us consider the 1d case (other cases can be easily derived). Let us denote by $\mathbf{X} \in \mathbb{R}^{C\times L}$ an input tensor of a convolutional layer, where $C$ is the number of channels and $L$ is the dimension of each feature map (a.k.a. spatial dimension). We denote $\mathbf{W} \in  \mathbb{R}^{D\times C\times S}$ the corresponding weight tensor, where $D$ is the number of output channels, $S$ the kernel size, and $\mathbf{Y} \in  \mathbb{R}^{D\times L}$ the output tensor. Disregarding padding (i.e. border effects), the convolution operation is defined through Equation~(\ref{convolution_operation}) and depicted in Figure~\ref{fig:convolution}, panel (1).

\begin{equation}
    y_{d,\ell}=\sum_{c=1}^C\sum_{\ell^\prime=1}^{S}x_{c,\ell+\ell^\prime-\lceil S/2\rceil}w_{d,c,\ell^\prime},\forall d, \ell\;.
    \label{convolution_operation}
\end{equation}

A shift layer is obtained when connections in $\mathbf{W}$ are pruned so that exactly one connection remains for each slice $\mathbf{W}_{d,c,\cdot},\forall d,c$. We denote $\ell_{d,c}$ the index such that $w_{d,c,\ell_{d,c}}$ is not pruned. Then the shift operation is defined in Equation~(\ref{shift_operation}) and depicted in Figure~\ref{fig:convolution}: panel (3).

\begin{eqnarray}
    y_{d,\ell}&=&\sum_{c=1}^Cx_{c,\ell + \ell_{d,c}-\lceil S/2\rceil}w_{d,c,\ell_{d,c}}\\
    &=& \sum_{c=1}^C\tilde{x}_{c,\ell}\tilde{w}_{d,c}\;,
    \label{shift_operation}
\end{eqnarray}
where $\tilde{x}_{c,\ell} = x_{c,\ell+\ell_{d,c}-\lceil S/2\rceil}$ and $\tilde{w}_{d,c} = w_{d,c,\ell_{d,c}}$.
We observe that the shift operation boils down to shifting the input tensor $\mathbf{X}$ then convoluting with a kernel of size 1 (in the equation, the kernel $\mathbf{\tilde{W}}$ is indexed only by the input and output feature maps). In the original work~\cite{wu2017shift}, the authors proposed to arbitrarily predetermine which shifts are used before training the architecture. To the contrary, in this work, we propose a method that learns which shifts to perform during the training process.

\subsection{Shift Attention Layer (SAL)}

The idea we propose is to enrich a classical convolution layer with a selective tensor $\mathbf{A}$ which aims at identifying which connections should be kept in each slice of the weight tensor. As such, we introduce $\mathbf{A} \in  \mathbb{R}^{D\times C\times L}$ a tensor containing as many elements as weights in the weight tensor. Each value of $\mathbf{A}$ is normalized between 0 and 1 and represents how important the corresponding weight in $\mathbf{W}$ is (c.f. Figure~\ref{fig:convolution}: panel (2)) with respect to the task to be solved. At the end of the training process, $\mathbf{A}$ becomes binary, with only one nonzero element per slice $\mathbf{A}_{d,c,\cdot}$, corresponding to the weights in $\mathbf{W}$ that should be kept.

More precisely, each slice $\mathbf{A}_{d,c,\cdot}$ is normalized using a softmax function with temperature $T$. The temperature is decreased smoothly along the training process. Note that in order to force the mask $\mathbf{A}$ to be selective, we first normalize each slice $\mathbf{A}_{d,c,\cdot}$ so that it has 1 standard deviation ($sd$).
Algorithm~\ref{algo:one} summarizes the training process of one layer. At the end of training, the selected weight in each slice $\mathbf{W}_{d,c,\cdot}$ corresponds to the maximum value in $\mathbf{A}_{d,c,\cdot}$.

\begin{algorithm}
\caption{SAL algorithm of one layer}
\textbf{Inputs}: Input tensor $\mathbf{X}$,\\ Initial softmax temperature $T$, Constant $\alpha < 1$.\\
\begin{algorithmic}
\FOR{ each training iteration}
\STATE $T \leftarrow \alpha T$
\FOR{$d:=1$ to $D$}
\FOR{$c:=1$ to $C$}
\STATE $\mathbf{A}_{d,c,\cdot} \leftarrow {\mathbf{A}_{d,c,\cdot} \over sd(\mathbf{A}_{d,c,\cdot})}$

\STATE $ \mathbf{A}_{d,c,\cdot} \leftarrow Softmax(T \mathbf{A}_{d,c,\cdot})$

\ENDFOR
\ENDFOR

\STATE $\mathbf{W}_A \leftarrow \mathbf{W} \cdot \mathbf{A}$ \text{($\cdot$ is the pointwise multiplication)}
\STATE Compute standard convolution as described in Equation~\ref{convolution_operation} using input tensor $\mathbf{X}$ and weight tensor $\mathbf{W}_A$ instead of $\mathbf{W}$ .
\STATE  Update $\mathbf{W}$ and $\mathbf{A}$ via back-propagation.

\ENDFOR
\end{algorithmic}
\label{algo:one}
\end{algorithm}

Note that contrary to the vanilla shift layers where the same number of shifts is performed in every direction, at the end of the training process the resulting shift layer can have an uneven distribution of shifts (c.f. Section~\ref{sec:experiments}). This comes with a price, which corresponds to the memory needed to retain which shift is performed for each input feature map.
In order to be fair, we thus reduce the number of feature maps in the networks we use in the experiments so that the total memory is comparable. In the next section we present related works, thus our proposed method can easily be compared to previously introduced methods.  

\section{Related Work}
\label{sec:relatedwork}
Let us introduce related works aiming at reducing complexity and memory footprints of CNNs.

In a first line of works, authors have proposed to binarize weights and activations~\cite{courbariaux2015binaryconnect, courbariaux2016binarized, rastegari2016xnor}, and then replace all multiplications by low cost multiplexers. Other works have proposed to use $K$-means to quantize weights~\cite{han2015deep, ullrich2017soft, wu2018deep} and thus reduce the model size of CNNs.

Another line of work relies on network pruning~\cite{han2015learning}. For example Li et al~\cite{li2016pruning} use the sum of absolute weights of each channel to select and prune redundant ones. In the same vein, He et al~\cite{he2018soft} proposed a novel Soft Filter Pruning (SFP) approach to prune dynamically filters in a soft manner. Using the reconstruction error of the last layer before classification, Yu et al~\cite{yu2018nisp} introduce the Neuron Importance Score Propagation (NISP) method. The main idea is to estimate the importance of each neuron in each layer in the backward propagation of the scores obtained from the reconstructed error. Huang et al~\cite{huang2018learning} propose a reinforcement learning based method, in which an agent is trained to maximize a reward function in order to improve the accuracy when pruning selected channels. Yamamoto et al~\cite{yamamoto2018pcas} use a channel-pruning technique based on attention statistics by adding attention blocks to each layer and updating them during the training process to evaluate the importance of each channel. Compared to these works, the method we propose can be seen as a structured pruning technique in which most of the weights of a convolution are pruned but one per slice.

For mobile applications with limited resources, specific neural network architectures have been introduced. Iandola et al~\cite{iandola2016squeezenet} propose a fire module to define SqueezeNet, a very small neural network with an acceptable accuracy. In~\cite{howard2017mobilenets,sandler2018mobilenetv2}, the authors use depthwise separable convolutions to define MobileNet and build lightweight CNNs. 

Other approaches focus on a decomposition of the convolution. Simonyan et al~\cite{simonyan2014very} reduce the number of parameters of VGG by replacing a $5\times5$ convolutions by two $3\times3$ convolutions each. In~\cite{szegedy2016rethinking}, the authors decompose $7\times7$ convolutions into $1\times7$ and $7\times1$ convolutions. Our proposed work builds upon Wu et al~\cite{wu2017shift}, where the authors introduce a shift operation, that we call shift layers throughout this article, as an alternative to spatial convolutions. The authors deconstruct $3\times3$ spatial convolutions into the concatenation of a shift transform on the input and $1\times1$ convolutions. As such, they are able to considerably reduce the complexity and memory usage of CNNs. In~\cite{hacene2018quantized}, the authors show it is possible to exploit this procedure for specific hardware targets. In~\cite{jeon2018constructing}, the authors propose an active shift layer (ASL) that improves the accuracy of the shift operation method, and replaces the fixed handcrafted integer parameters of the shift layers by learned floating point parameters. Consequently, the result architecture requires to perform interpolations and does not fall into the original shift layer formulation.

In this contribution, we introduce Shift Attention Layers (SALs), a novel pruning-shift operation method that ends up replacing convolutions with shift layers by the end of training. Contrary to~\cite{jeon2018constructing}, the proposed methodology result in a network that is exactly as described in~\cite{wu2017shift}, with the main difference that shifts are learned instead of being predetermined. We demonstrate in this paper that it results in better accuracy for the exact same complexity and number of parameters.

\section{Experiments}
\label{sec:experiments}
In this section we present the benchmark protocol used to evaluate the proposed method, and then compare the obtained performance with a CNN baseline, pruning methods and vanilla shift layers.

\subsection{Benchmark Protocol}
We evaluate using three object recognition datasets: CIFAR10, CIFAR100 and ImageNet ILSVRC 2012.
To be comparable with the relevant literature, we use the same architectures from previous papers~\cite{yamamoto2018pcas,wu2017shift,jeon2018constructing}, that is Resnet-20/56 on CIFAR10 and Resnet-20/50 on CIFAR100 using the following parameters: the training process consists of $300$ epochs for Resnet-20 and $400$ epochs for Resnet-50/56, the learning rate starts at $0.1$ and is divided by $10$ after each $100$ epochs. 

For the evaluations using CIFAR10 or CIFAR100~\cite{he2016identity}, the training batch size is $128$, the initial/final softmax temperatures are $6.7$/$0.02$, and the temperature is multiplied at each step by $\alpha = 0.99994,0.99996$ when using $300,400$ epochs respectively.  
For ImageNet ILSVRC 2012, we used Resnet-w32 and Resnet-w64 defined in~\cite{jeon2018constructing} using the following parameters: the total number of epochs is $90$, the training batch size is $1024$, the learning rate starts at $0.1$ and is divided by $10$ after each $30$ epochs, initial/final softmax temperatures are $6.7$/$0.016$ so that the temperature update at each step is $\alpha = 0.99995$. We also used standard data augmentation defined in~\cite{krizhevsky2012imagenet}. 

Let us point out that the values of temperatures were obtained by using a grid search. The fact the final temperature is not zero means that the tensors $\mathbf{A}$ may contain nonbinary values. This is why we binarize $\mathbf{A}$ using a hard max to obtain the corresponding shift layers before evaluating on the test set.


\subsection{Results}

The first experiment consists in comparing the accuracy, memory and number of operations of the proposed method compared to other shift layer based methods and pruning methods.
We run tests using CIFAR10 and CIFAR100. Table~\ref{table:shift} shows that our method achieves a better accuracy with fewer parameters than the baseline and other shift-module based methods. Tables~\ref{table:pruning10} and~\ref{table:pruning100} show that the proposed method is comparable or better in term of accuracy and number of parameters/floating point operations (FLOPs) with other pruning methods.

\begin{table*}
\centering

    \caption{Comparison of accuracy and number of parameters between the baseline architecture (ResNet20), ShiftNet, ASNet, and SANet (the proposed method) on CIFAR10 and CIFAR100.}
    \centering
    {\renewcommand{\arraystretch}{1.3}%
    \begin{tabular} { | c || c | c | c | c | c |} 

  \cline{3-6}
    \multicolumn{2}{l|}{} &  \multicolumn{2}{c|}{CIFAR10} & \multicolumn{2}{c|}{CIFAR100} \\
  \cline{3-6}
  \multicolumn{2}{l|}{} & Accuracy & Params (M) &  Accuracy & Params (M) \\
  
   \hline
  CLs & Baseline & $94.66\%$ & $1.22$ & $73.7\%$ & $1.24$ \\
  \hline
  \hline
  SLs & ShiftNet~\cite{wu2017shift}   & $93.17\%$  & $1.2$ &  $72.56\%$ & $1.23$ \\
   \cline{2-6}
   
   & SANet (ours)   & $\mathbf{95.52}\%$  & $\mathbf{0.98} $ &  $\mathbf{77.39}\%$ & $\mathbf{1.01}$ \\
   \hline
   \hline
   Interpolate & ASNet~\cite{jeon2018constructing}   & $94.53\%$  & $0.99$ &  $76.73\%$ & $1.02$ \\
   \hline
   
    \end{tabular}} \quad
    \vspace{.3cm}
  
  \label{table:shift}
  \vspace{-.5cm}
  \end{table*}

  \begin{table*}
\centering

    \caption{Comparison of accuracy, number of parameters and number of floating point operations (FLOPs) between baseline architecture (Resnet-56), SANet (the proposed method) , and some other pruning methods on CIFAR10. Note that the number between () refers to the result obtained by the baseline used for each method.}
    \centering
    {\renewcommand{\arraystretch}{1.3}%
    \begin{tabular} { | c || c | c | c | c |} 

  \cline{3-5}
    \multicolumn{2}{l|}{} &  \multicolumn{3}{c|}{CIFAR10}  \\
  \cline{3-5}
  \multicolumn{2}{l|}{} & Accuracy & Params (M) & FLOPs (M)  \\
  
   \hline
   &Pruned-B~\cite{li2016pruning}   & $93.06\% (93.04)$  & $0.73 (0.85)$ &  $91 (126)$ \\
   \cline{2-5}
Pruning  & NISP~\cite{yu2018nisp}   & $93.01\% (93.04)$  & $0.49 (0.85)$ &  $71 (126)$ \\
   \cline{2-5}
  & PCAS~\cite{yamamoto2018pcas}   & $93.58\% (93.04)$  & $0.39 (0.85)$ &  $56 (126)$ \\
   \cline{2-5}
 & SANet (ours)   & $\mathbf{94}\% (93.04)$  & $\mathbf{0.36} (0.85) $ &  $\mathbf{42} (126)$ \\
    
   \hline
   
    \end{tabular}} \quad
    \vspace{.3cm}
  
  \label{table:pruning10}
  \vspace{-.5cm}
  \end{table*}

    \begin{table*}
\centering

    \caption{Comparison of accuracy, number of parameters and number of floating point operations (FLOPs) between baseline architecture (Resnet-50), SANet (the proposed method) , and some other pruning methods on CIFAR100. Note that the number between () refers to the result obtained by the baseline used for each method.}
    \centering
    {\renewcommand{\arraystretch}{1.3}%
    \begin{tabular} { | c || c | c | c | c |} 

  \cline{3-5}
    \multicolumn{2}{l|}{} &  \multicolumn{3}{c|}{CIFAR100} \\
  \cline{3-5}
  \multicolumn{2}{l|}{} &  Accuracy & Params (M) & FLOPs (M) \\
  
   \hline
  &Pruned-B~\cite{li2016pruning}   & $73.6\% (74.46)$ & $7.83 (17.1)$ & $616 (1409)$ \\
   \cline{2-5}
  Pruning &PCAS~\cite{yamamoto2018pcas}  & $73.84\% (74.46)$ & $4.02 (17.1)$ & $475 (1409)$ \\
   \cline{2-5}
  &SANet (ours)   & $\mathbf{77.6}\% (78) $ &  $\mathbf{3.9}$ $(16.9)$ & $\mathbf{251} (1308)$ \\
    
   \hline
   
    \end{tabular}} \quad
    \vspace{.3cm}
  
  \label{table:pruning100}
  \vspace{-.1cm}
  \end{table*}

As mentioned in Section~\ref{sec:methodology}, the number of shifts in each direction can be uneven at the end of the training procedure. As such, in a second run of experiments, we look at the proportions of each shifts (relative position in kernel) depending on the layer depth in the architecture. In Figure~\ref{fig:HM20}, a heat-map represents the distribution of kept weights at different relative positions through $\mathbf{W}_{d,c,\cdot,\cdot}, \forall d,c$ slices, for the $4$ first layers (first row), and the $4$ last layers (second row), of Resnet-20 trained on CIFAR10, where attention tensors $\mathbf{A}$ values are initially drawn uniformly at random. We observe that at the end of the training process, first layers seem to yield a uniform distribution of kept weights. To the contrary, for the last layers, there is a clear asymmetry that favours corners. This interestingly suggests that shift-layers would benefit from a non regular number of shifts in each direction. 

To see how much the initialization of $\mathbf{A}$ is related to the distribution of kept weights, we then perform another experiment where $\mathbf{A}$ is initialized uniformly at random but then the centre value $\mathbf{A}_{d,c,\lfloor S/2 \rfloor,\lfloor S/2 \rfloor}$ is changed to the maximum over the corresponding slice $\max(\mathbf{A}_{d,c,\cdot,\cdot})$. We observe in Figure~\ref{fig:HM20C} that almost all kept weights in the first layer are slice centres. Subsequent layers yield an almost uniform distribution, and we observe the same kept weights distribution for the last layers as in the previous experiment. We also plot a heat-map of kept weights in Resnet-56 trained on CIFAR10, and where attention tensor $\mathbf{A}$ values are initially drawn uniformly at random. Figure~\ref{fig:HM56} shows that for the first layers, the number of kept weights is more important in the centre row than at other positions. However, we see on the last layers that there is more kept weights in the corners than other positions. 

\begin{figure*}
    \centering
    \begin{tikzpicture}
        
       \foreach \x in {0,3,6,9}{\foreach \y in {0,-3}{
    \draw (0+\x,0+\y)--(2+\x,0+\y)--(2+\x,-2+\y)--(0+\x,-2+\y)--cycle;
    \draw (0+\x,-0.4+\y)--(2+\x,-0.4+\y);
    \draw (0+\x,-0.8+\y)--(2+\x,-0.8+\y);
    \draw (0+\x,-1.2+\y)--(2+\x,-1.2+\y);
    \draw (0+\x,-1.6+\y)--(2+\x,-1.6+\y);
    \draw (0.4+\x,0+\y)--(.4+\x,-2+\y);
    \draw (0.8+\x,0+\y)--(.8+\x,-2+\y);
    \draw (1.2+\x,0+\y)--(1.2+\x,-2+\y);
    \draw (1.6+\x,0+\y)--(1.6+\x,-2+\y);

    }}
\def\c{{0.27266917, 0.17452117, 0.27993939, 0.18906162, 0.36718205,
      0.3062008 , 0.27457231, 0.32377218, 0.28862942, 0.29917225,
       00.26189547, 0.19121838, 0.33964026, 0.24069234, 0.22302307,
       0.27949895, 0.32507821, 0.23391968, 0.31105382, 0.20937699
}}

    \def\x{{0,0.4,0.8,1.2,1.6,3,3.4,3.8,4.2,4.6,6,6.4,6.8,7.2,7.6,9,9.4,9.8,10.2,10.6}}
    \def \y{0}
    \foreach \i in {0,...,19}{
    
    \draw [fill opacity=\c[\i), fill=black] (0+\x[\i],0+\y)--(.4+\x[\i],0+\y)--(.4+\x[\i],-.4+\y)--(0+\x[\i],-.4+\y)--cycle;
    }

    \def\c{{0.27266917, 0.27630428, 0.45442471, 0.27266917, 0.26539895,
       0.232401  , 0.232401  , 0.43622902, 0.25700093, 0.27808659 ,
       0.25482776, 0.28663245, 0.54460382, 0.491596  , 0.30430172,
       0.12873675, 0.23041358, 0.45830992, 0.27949895, 0.18834041
}}
\def \y{-0.4}
    \foreach \i in {0,...,19}{
    
    \draw [fill opacity=\c[\i), fill=black] (0+\x[\i],0+\y)--(.4+\x[\i],0+\y)--(.4+\x[\i],-.4+\y)--(0+\x[\i],-.4+\y)--cycle;
    }

    \def\c{{0.42534383, 0.43624916, 0.85428692, 0.51985671, 0.4180736,
       0.3020077251, 0.46082895, 0.79117092, 0.49245744, 0.35188639,
       0.34317412, 0.33964026, 0.75663508, 0.48806215, 0.33257256,
       0.15678553, 0.46181602, 0.88254772, 0.59855382, 0.41623675
}}
\def \y{-0.8}
    \foreach \i in {0,...,19}{
    
    \draw [fill opacity=\c[\i), fill=black] (0+\x[\i],0+\y)--(.4+\x[\i],0+\y)--(.4+\x[\i],-.4+\y)--(0+\x[\i],-.4+\y)--cycle;
    
    }
    
    \def\c{{0.23995317, 0.26539895, 0.5162216 , 0.27266917, 0.37445227 ,
       0.28160087, 0.36945777, 0.5276002 , 0.3062008 , 0.32728646,
       0.30430172, 0.26189547, 0.51633298, 0.28309859, 0.32197099,
       0.31806602, 0.37416358, 0.53895016, 0.37416358, 0.37065748
}}
\def \y{-1.2}
    \foreach \i in {0,...,19}{
    
    \draw [fill opacity=\c[\i), fill=black] (0+\x[\i],0+\y)--(.4+\x[\i],0+\y)--(.4+\x[\i],-.4+\y)--(0+\x[\i],-.4+\y)--cycle;
    }
  \def\c{{0.3162905 , 0.32356072, 0.32719583, 0.11272429, 0.33810116,
       0.37297205, 0.16914403, 0.40811481, 0.32377218, 0.25348666,
       0.34670797, 0.27249703, 0.42091891, 0.26189547, 0.17708296,
       0.31105382, 0.28651114, 0.31806602, 0.26898065, 0.24443797
}}  
    \def \y{-1.6}
    \foreach \i in {0,...,19}{
    
    \draw [fill opacity=\c[\i), fill=black] (0+\x[\i],0+\y)--(.4+\x[\i],0+\y)--(.4+\x[\i],-.4+\y)--(0+\x[\i],-.4+\y)--cycle;
    }
    
    \def\c{{0.39627183, 0.33491112, 0.36834061, 0.3725193 , 0.3931928,
       0.40514019, 0.34532556, 0.32626375, 0.3523368 , 0.39308962,
       0.45043236, 0.33295587, 0.4636073 , 0.38807101, 0.47788014,
       0.8976743 , 0.4104281 , 0.49273321, 0.41898784, 0.77542379}}
    \def \y{-3}
    \foreach \i in {0,...,19}{
    
    \draw [fill opacity=\c[\i), fill=black] (0+\x[\i],0+\y)--(.4+\x[\i],0+\y)--(.4+\x[\i],-.4+\y)--(0+\x[\i],-.4+\y)--cycle;
     }

    \def\c{{0.32809327, 0.27508993, 0.27618959, 0.32039569, 0.33315168,
       0.31399408, 0.25680866, 0.29756149, 0.26776372, 0.32976937,
       0.27608407, 0.21943185, 0.35908615, 0.3263684 , 0.27959738,
       0.40099045, 0.05201682, 0.06716096, 0.03160516, 0.35446064
}}
\def \y{-3.4}
    \foreach \i in {0,...,19}{
    
    \draw [fill opacity=\c[\i), fill=black] (0+\x[\i],0+\y)--(.4+\x[\i],0+\y)--(.4+\x[\i],-.4+\y)--(0+\x[\i],-.4+\y)--cycle;
    }
    
    \def\c{{0.32897299, 0.28256758, 0.29114488, 0.31357783, 0.34216884,
       0.30654464, 0.29602778, 1.        , 0.27543227, 0.30610644,
       0.24556214, 0.21240522, 0.39290182, 0.32285508, 0.28003655,
       0.33690221, 0.02063115, 0.25064647, 0.        , 0.31890483
}}
\def \y{-3.8}
    \foreach \i in {0,...,19}{
    
    \draw [fill opacity=\c[\i), fill=black] (0+\x[\i],0+\y)--(.4+\x[\i],0+\y)--(.4+\x[\i],-.4+\y)--(0+\x[\i],-.4+\y)--cycle;
    }
    
    \def\c{{0.35602448, 0.32457437, 0.29642322, 0.29334419, 0.3661413,
       0.28770194, 0.281348  , 0.29339857, 0.25352215, 0.31925251,
       00.24687963, 0.22470182, 0.39575639, 0.35557284, 0.29935978,
       0.35731388, 0.03292204, 0.06277136, 0.01602206, 0.31275938
}}
\def \y{-4.2}
    \foreach \i in {0,...,19}{
    
    \draw [fill opacity=\c[\i), fill=black] (0+\x[\i],0+\y)--(.4+\x[\i],0+\y)--(.4+\x[\i],-.4+\y)--(0+\x[\i],-.4+\y)--cycle;
    }
  \def\c{{0.41694533, 0.38703473, 0.39077356, 0.36064303, 0.4167254,
       0.35912893, 0.30764015, 0.29361767, 0.30040981, 0.39703344,
       0.43418328, 0.31429138, 0.47129268, 0.39114516, 0.40475926 ,
       0.86497174, 0.42184108, 0.43325405, 0.422719  , 0.812077
}}  
    \def \y{-4.6}
    \foreach \i in {0,...,19}{
    
    \draw [fill opacity=\c[\i), fill=black] (0+\x[\i],0+\y)--(.4+\x[\i],0+\y)--(.4+\x[\i],-.4+\y)--(0+\x[\i],-.4+\y)--cycle;
    }
    \def \xx{-3.1}
    
     \foreach \i in {0,...,10}{
    \draw[fill opacity=\i*.1, fill=black](15+\xx,-3.8+.4*\i-1)--(15+0.4+\xx,-3.8+.4*\i-1)--(15+0.4+\xx,-3.8+.4*\i +.4-1)--(15+\xx,-3.8+.4*\i+.4-1)--cycle;
    
    }
    \node at (16+\xx,-3.6-1){$0.0258$};
    \node at (16+\xx,.4-1){$0.0673$};
    
    \foreach \i in{0,...,3}{
    \foreach \j in{0,-3}{
    \node at (0+\i*3,.2+\j){$-2$};
    \node at (0.5+\i*3,.2+\j){$-1$};
    \node at (1+\i*3,.2+\j){$0$};
    \node at (1.4+\i*3,.2+\j){$1$};
    \node at (1.8+\i*3,.2+\j){$2$};

     \node at (-.3+\i*3,-.2+\j){$-2$};
     \node at (-.3+\i*3,-.6+\j){$-1$};
     \node at (-.2+\i*3,-1+\j){$0$};
     \node at (-.2+\i*3,-1.4+\j){$1$};
     \node at (-.2+\i*3,-1.8+\j){$2$};}}

    \end{tikzpicture}
    \caption{Heat maps representing the average values in $\mathbf{A}$ for various layers in the Resnet-20 architecture with $5 \times 5$ kernels trained on CIFAR10. In this experiment, values in $\mathbf{A}$ are initialized uniformly at random. The first row represents the $4$ first layers and the second row the $4$ last layers.}
    \label{fig:resnet20_5X5}
\end{figure*}
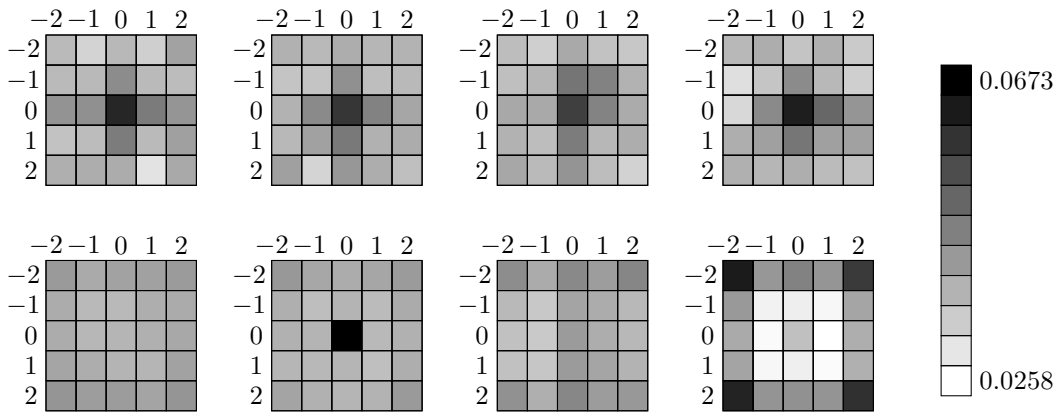

For further results, we run an experiment in which we replace all $3\times 3$ Resnet-20 kernels by $5\times 5$ kernels, and train the network on CIFAR10. We observe in Figure~\ref{fig:resnet20_5X5} that the weights of the center in first layers are more important that other positions. We also see that on the last layers the weight distribution is still not uniform, and the weights on the corners are more important in the last layer. 

From all these experiments, we consistently observe that in deeper layers, the method tends to keep more weights in corner positions than others, and this independently from initialization process or neural network architecture. This finding interestingly questions the hyperparameters used by the corresponding architectures. It clearly seems the network is more interested in locality in the initial layers than it is in the last layers. Based on this finding, we modified the vanilla shift layer method, using an equivalent uneven distribution of shifts as the one found in our experiments. As such, shifts are predetermined but not uniform. We obtained an accuracy of $94.8\%$ on Resnet-20 and CIFAR10, to be compared to the $93.17\%$ accuracy from Table~\ref{table:shift}. Interestingly, this accuracy is even better than the results obtained using the method in~\cite{jeon2018constructing}. On the other hand, the obtained accuracy remains lower than that of the proposed method, suggesting that selecting the shifts during the learning process is still more efficient than having a good choice of predetermined shift proportions.
  

\begin{figure*}
 \begin{center}

  \begin{tikzpicture}[thick, scale=1]
    \foreach \x in {0,3,6,9}{\foreach \y in {0,-2}{
    \draw (0+\x,0+\y)--(1.2+\x,0+\y)--(1.2+\x,-1.2+\y)--(0+\x,-1.2+\y)--cycle;
    \draw (0+\x,-0.4+\y)--(1.2+\x,-0.4+\y);
    \draw (0+\x,-0.8+\y)--(1.2+\x,-0.8+\y);
    \draw (0.4+\x,0+\y)--(.4+\x,-1.2+\y);
    \draw (0.8+\x,0+\y)--(.8+\x,-1.2+\y);
    \node[text width=3cm] at (1.4+\x,.2+\y) 
    {$-1$};
    \node[text width=3cm] at (2+\x,.2+\y) 
    {$0$};
    \node[text width=3cm] at (2.4+\x,.2+\y) 
    {$1$};
    
    \node[text width=3cm] at (1+\x,-.2+\y) 
    {$-1$};
    \node[text width=3cm] at (1.25+\x,-.6+\y) 
    {$0$};
    \node[text width=3cm] at (1.3+\x,-1+\y) 
    {$1$};

    }}
    
    \def\y{0}
    
    \def\x{{0,0.4,0.8,3,3.4,3.8,6,6.4,6.8,9,9.4,9.8}}
    \def\c{{0.17323136, 0.14225621, 0.18929254, 0.1957935 , 0.16405354,
       0.18087954, 0.19847036, 0.16634799, 0.17629063, 0.1709369 ,
       0.14646272, 0.16978967
}}
    \foreach \i in {0,...,11}{
    \draw [fill opacity=\c[\i), fill=black] (0+\x[\i],0+\y)--(.4+\x[\i],0+\y)--(.4+\x[\i],-.4+\y)--(0+\x[\i],-.4+\y)--cycle;
    }
    
    \def\y{-0.4}

    \def\c{{0.16290631, 0.20152964, 0.19655832, 0.1499044 , 0.17437859,
       0.15258126, 0.18087954, 0.15755258, 0.13690249, 0.1376673 ,
       0.18470363, 0.16826004
}}
    \foreach \i in {0,...,11}{
    
    \draw [fill opacity=\c[\i), fill=black] (0+\x[\i],0+\y)--(.4+\x[\i],0+\y)--(.4+\x[\i],-.4+\y)--(0+\x[\i],-.4+\y)--cycle;
    }
    
    \def\y{-0.8}

    \def\c{{0.16902486, 0.13728489, 0.15296367, 0.1418738 , 0.18546845,
       0.18011472, 0.18240918, 0.16749522, 0.15793499, 0.16328872,
       0.18852772, 0.19541109
}}
    \foreach \i in {0,...,11}{
    
    \draw [fill opacity=\c[\i), fill=black] (0+\x[\i],0+\y)--(.4+\x[\i],0+\y)--(.4+\x[\i],-.4+\y)--(0+\x[\i],-.4+\y)--cycle;
    }

     \def\y{-2}
    
    \def\x{{0,0.4,0.8,3,3.4,3.8,6,6.4,6.8,9,9.4,9.8}}
    \def\c{{0.20382409, 0.15066922, 0.208413  , 0.20764818, 0.16405354,
       0.20611855, 0.19694073, 0.17476099, 0.20382409, 0.29254302,
       0.1001912 , 0.28910134
}}
    \foreach \i in {0,...,11}{
    \draw [fill opacity=\c[\i), fill=black] (0+\x[\i],0+\y)--(.4+\x[\i],0+\y)--(.4+\x[\i],-.4+\y)--(0+\x[\i],-.4+\y)--cycle;
    }
    
    \def\y{-2.4}

    \def\c{{0.14722753, 0.08871893, 0.14837476, 0.14531549, 0.10975143,
       0.14684512, 0.13499044, 0.11548757, 0.13652008, 0.08565966,
       0.        , 0.07036329
}}
    \foreach \i in {0,...,11}{
    
    \draw [fill opacity=\c[\i), fill=black] (0+\x[\i],0+\y)--(.4+\x[\i],0+\y)--(.4+\x[\i],-.4+\y)--(0+\x[\i],-.4+\y)--cycle;
    }
    
    \def\y{-2.8}

    \def\c{{0.21223709, 0.15449331, 0.21070746, 0.19349904, 0.15181644,
       0.20038241, 0.19311663, 0.17017208, 0.19885277, 0.29483748,
       0.10630975, 0.28642447
}}
    \foreach \i in {0,...,11}{
    
    \draw [fill opacity=\c[\i), fill=black] (0+\x[\i],0+\y)--(.4+\x[\i],0+\y)--(.4+\x[\i],-.4+\y)--(0+\x[\i],-.4+\y)--cycle;
    
    }
    \def \nx{-3.5}
    \foreach \i in {0,...,10}{
    \draw[fill opacity=\i*.1, fill=black](15+\nx,-3.8+.4*\i)--(15+0.4+\nx,-3.8+.4*\i)--(15+0.4+\nx,-3.8+.4*\i +.4)--(15+\nx,-3.8+.4*\i+.4)--cycle;
     }
    \node at (16+\nx,-3.6){$0.0604$};
    \node at (16+\nx,.4){$0.3172$};
    \end{tikzpicture}
   \end{center}
   
    \caption{Heat maps representing the average values in $\mathbf{A}$ for various layers in the Resnet-20 architecture trained on CIFAR10. In this experiment, values in $\mathbf{A}$ are initialized uniformly at random. The first row represents the $4$ first layers and the second row the $4$ last layers of Resnet-20.}
    \label{fig:HM20}
\end{figure*}
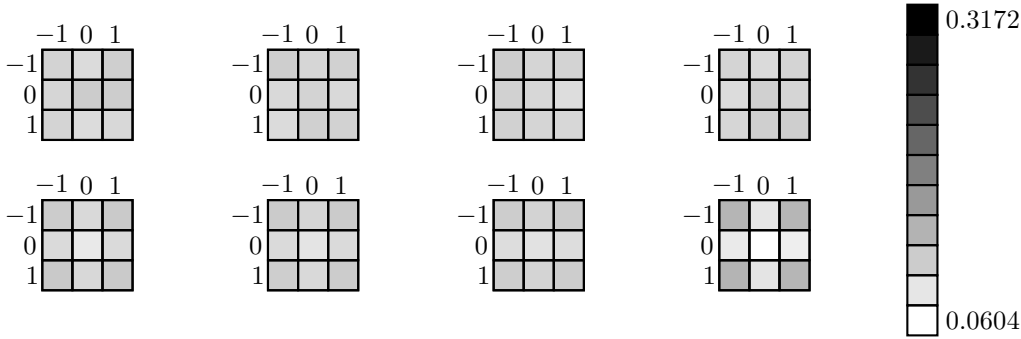

\begin{figure*}
\begin{center}

\begin{tikzpicture}[thick, scale=1]
    \foreach \x in {0,3,6,9}{\foreach \y in {0,-2}{
    \draw (0+\x,0+\y)--(1.2+\x,0+\y)--(1.2+\x,-1.2+\y)--(0+\x,-1.2+\y)--cycle;
    \draw (0+\x,-0.4+\y)--(1.2+\x,-0.4+\y);
    \draw (0+\x,-0.8+\y)--(1.2+\x,-0.8+\y);
    \draw (0.4+\x,0+\y)--(.4+\x,-1.2+\y);
    \draw (0.8+\x,0+\y)--(.8+\x,-1.2+\y);
\node[text width=3cm] at (1.4+\x,.2+\y) 
{$-1$};
    \node[text width=3cm] at (2+\x,.2+\y) 
    {$0$};
    \node[text width=3cm] at (2.4+\x,.2+\y) 
    {$1$};
    
    \node[text width=3cm] at (1+\x,-.2+\y) 
    {$-1$};
    \node[text width=3cm] at (1.25+\x,-.6+\y) 
    {$0$};
    \node[text width=3cm] at (1.3+\x,-1+\y) 
    {$1$};

    }}
    
    \def\y{0}
    
    \def\x{{0,0.4,0.8,3,3.4,3.8,6,6.4,6.8,9,9.4,9.8}}
    \def\c{{0.05315488, 0.03135755, 0.04627151, 0.15066922, 0.12772467,
       0.15219885, 0.14799235, 0.13804971, 0.16749522, 0.15869981,
       0.15602294, 0.16711281
}}
    \foreach \i in {0,...,11}{
    \draw [fill opacity=\c[\i), fill=black] (0+\x[\i],0+\y)--(.4+\x[\i],0+\y)--(.4+\x[\i],-.4+\y)--(0+\x[\i],-.4+\y)--cycle;}
    
    \def\y{-0.4}

    \def\c{{0.08565966, 1.        , 0.0833652 , 0.14722753, 0.31128107,
       0.15181644, 0.14302103, 0.33193117, 0.14760994, 0.15984704,
       0.2210325 , 0.1793499
}}
    \foreach \i in {0,...,11}{
    
    \draw [fill opacity=\c[\i), fill=black] (0+\x[\i],0+\y)--(.4+\x[\i],0+\y)--(.4+\x[\i],-.4+\y)--(0+\x[\i],-.4+\y)--cycle;}
    
    \def\y{-0.8}

    \def\c{{0.0707457 , 0.07418738, 0.07992352, 0.17552581, 0.15946463,
       0.14952199, 0.13804971, 0.16137667, 0.14913958, 0.14837476,
       0.15219885, 0.18279159
}}
    \foreach \i in {0,...,11}{
    
    \draw [fill opacity=\c[\i), fill=black] (0+\x[\i],0+\y)--(.4+\x[\i],0+\y)--(.4+\x[\i],-.4+\y)--(0+\x[\i],-.4+\y)--cycle;}

     \def\y{-2}
    
    \def\x{{0,0.4,0.8,3,3.4,3.8,6,6.4,6.8,9,9.4,9.8}}
    \def\c{{0.19196941, 0.15564054, 0.19655832, 0.19961759, 0.15908222,
       0.20573614, 0.19235182, 0.16175908, 0.20114723, 0.28298279,
       0.0959847 , 0.29292543
}}
    \foreach \i in {0,...,11}{
    \draw [fill opacity=\c[\i), fill=black] (0+\x[\i],0+\y)--(.4+\x[\i],0+\y)--(.4+\x[\i],-.4+\y)--(0+\x[\i],-.4+\y)--cycle;}
    
    \def\y{-2.4}

    \def\c{{0.1414914 , 0.13728489, 0.14263862, 0.14378585, 0.13499044,
       0.1418738 , 0.12237094, 0.16099426, 0.13499044, 0.06921606,
       0.05736138, 0.07151052
}}
    \foreach \i in {0,...,11}{
    
    \draw [fill opacity=\c[\i), fill=black] (0+\x[\i],0+\y)--(.4+\x[\i],0+\y)--(.4+\x[\i],-.4+\y)--(0+\x[\i],-.4+\y)--cycle;}
    
    \def\y{-2.8}

    \def\c{{0.19502868, 0.15678776, 0.20726577, 0.19082218, 0.14913958,
       0.2       , 0.18240918, 0.17131931, 0.19770554, 0.27839388,
       0.08680688, 0.29024857
}}
    \foreach \i in {0,...,11}{
    
    \draw [fill opacity=\c[\i), fill=black] (0+\x[\i],0+\y)--(.4+\x[\i],0+\y)--(.4+\x[\i],-.4+\y)--(0+\x[\i],-.4+\y)--cycle;}
    \def \nx{-3.5}
    \foreach \i in {0,...,10}{
    \draw[fill opacity=\i*.1, fill=black](15+\nx,-3.8+.4*\i)--(15+0.4+\nx,-3.8+.4*\i)--(15+0.4+\nx,-3.8+.4*\i +.4)--(15+\nx,-3.8+.4*\i+.4)--cycle;}
    \node at (16+\nx,-3.6){$0.0604$};
    \node at (16+\nx,.4){$0.3172$};
    
    \end{tikzpicture}
   \end{center}
   
    \caption{Heat maps representing the average values in $\mathbf{A}$ for various layers in the Resnet-20 architecture trained on CIFAR10. In this experiment, values in $\mathbf{A}$ are initialized uniformly at random but the centre value that takes the maximum over the corresponding slice. The first row represents the $4$ first layers and the second row the $4$ last layers of Resnet-20.}
    \label{fig:HM20C}
    \vspace{-.3cm}
\end{figure*}
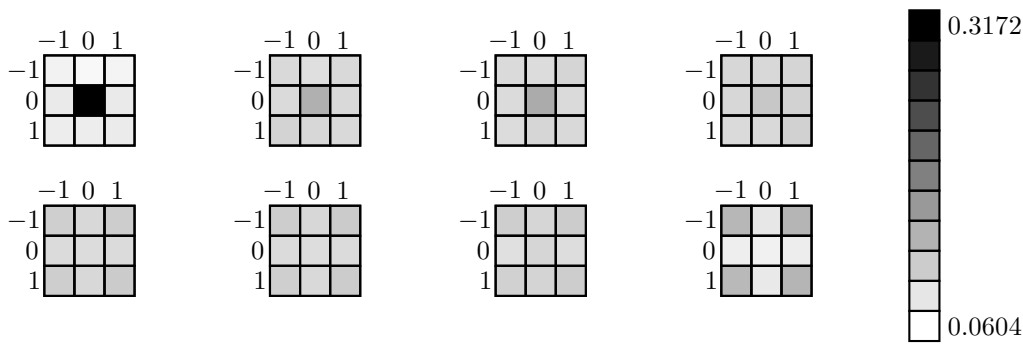


\begin{figure*}
\begin{center}

\begin{tikzpicture}[thick, scale=1]
    \foreach \x in {0,3,6,9}{\foreach \y in {0,-2}{
    \draw (0+\x,0+\y)--(1.2+\x,0+\y)--(1.2+\x,-1.2+\y)--(0+\x,-1.2+\y)--cycle;
    \draw (0+\x,-0.4+\y)--(1.2+\x,-0.4+\y);
    \draw (0+\x,-0.8+\y)--(1.2+\x,-0.8+\y);
    \draw (0.4+\x,0+\y)--(.4+\x,-1.2+\y);
    \draw (0.8+\x,0+\y)--(.8+\x,-1.2+\y);
    \node[text width=3cm] at (1.4+\x,.2+\y) 
    {$-1$};
    \node[text width=3cm] at (2+\x,.2+\y) 
    {$0$};
    \node[text width=3cm] at (2.4+\x,.2+\y) 
    {$1$};
    
    \node[text width=3cm] at (1+\x,-.2+\y) 
    {$-1$};
    \node[text width=3cm] at (1.25+\x,-.6+\y) 
    {$0$};
    \node[text width=3cm] at (1.3+\x,-1+\y) 
    {$1$};

    }}
    
    \def\y{0}
    
    \def\x{{0,0.4,0.8,3,3.4,3.8,6,6.4,6.8,9,9.4,9.8}}
    \def\c{{0.16861370716510904, 0.19042056074766353, 0.10786604361370716, 0.13356697819314645, 0.19392523364485986, 0.14641744548286606, 0.1974299065420561, 0.2059968847352025, 0.11098130841121498, 0.13746105919003113, 0.2542834890965732 , 0.14174454828660435
}}
    \foreach \i in {0,...,11}{
    \draw [fill opacity=\c[\i), fill=black] (0+\x[\i],0+\y)--(.4+\x[\i],0+\y)--(.4+\x[\i],-.4+\y)--(0+\x[\i],-.4+\y)--cycle;}
    
    \def\y{-0.4}

    \def\c{{0.61588683, 0.83460283, 0.6039173 , 0.63873776, 0.88139282,
       0.59085963, 0.67247008, 0.61153428, 0.69640914, 0.44396083,
       1.        , 0.67355822
}}
    \foreach \i in {0,...,11}{
    
    \draw [fill opacity=\c[\i), fill=black] (0+\x[\i],0+\y)--(.4+\x[\i],0+\y)--(.4+\x[\i],-.4+\y)--(0+\x[\i],-.4+\y)--cycle;}
    
    \def\y{-0.8}

    \def\c{{0.42219804, 0.51904244, 0.37323177, 0.21436344, 0.5179543 ,
       0.50598477, 0.50272035, 0.42981502, 0.32100109, 0.3710555 ,
       0.47986942, 0.2132753
}}
    \foreach \i in {0,...,11}{
    
    \draw [fill opacity=\c[\i), fill=black] (0+\x[\i],0+\y)--(.4+\x[\i],0+\y)--(.4+\x[\i],-.4+\y)--(0+\x[\i],-.4+\y)--cycle;}

     \def\y{-2}
    
    \def\x{{0,0.4,0.8,3,3.4,3.8,6,6.4,6.8,9,9.4,9.8}}
    \def\c{{0.61915125, 0.59412405, 0.60065288, 0.65070729, 0.49727965,
       0.65832427, 0.60065288, 0.54080522, 0.56800871, 0.84983678,
       0.38955386, 0.84983678
}}
    \foreach \i in {0,...,11}{
    \draw [fill opacity=\c[\i), fill=black] (0+\x[\i],0+\y)--(.4+\x[\i],0+\y)--(.4+\x[\i],-.4+\y)--(0+\x[\i],-.4+\y)--cycle;}
    
    \def\y{-2.4}

    \def\c{{0.35690968, 0.40043526, 0.32100109, 0.36452666, 0.26550598,
       0.390642  , 0.3101197 , 0.32317737, 0.34820457, 0.26224157,
       0.        , 0.28291621
}}
    \foreach \i in {0,...,11}{
    
    \draw [fill opacity=\c[\i), fill=black] (0+\x[\i],0+\y)--(.4+\x[\i],0+\y)--(.4+\x[\i],-.4+\y)--(0+\x[\i],-.4+\y)--cycle;}
    
    \def\y{-2.8}

    \def\c{{0.54733406, 0.55277476, 0.53427639, 0.61044614, 0.50163221,
       0.58650707, 0.61153428, 0.62023939, 0.60282916, 0.76605005,
       0.35690968, 0.76822633
}}
    \foreach \i in {0,...,11}{
    
    \draw [fill opacity=\c[\i), fill=black] (0+\x[\i],0+\y)--(.4+\x[\i],0+\y)--(.4+\x[\i],-.4+\y)--(0+\x[\i],-.4+\y)--cycle;}
    \def \nx{-3.5}
    \foreach \i in {0,...,10}{
    \draw[fill opacity=\i*.1, fill=black](15+\nx,-3.8+.4*\i)--(15+0.4+\nx,-3.8+.4*\i)--(15+0.4+\nx,-3.8+.4*\i +.4)--(15+\nx,-3.8+.4*\i+.4)--cycle;}
    \node at (16+\nx,-3.6){$0.0658$};
    \node at (16+\nx,.4){$0.1787$};
    
    \end{tikzpicture}
   \end{center}
   
    \caption{Heat maps representing the average values in $\mathbf{A}$ for various layers in the Resnet-56 architecture trained on CIFAR10. In this experiment, values in $\mathbf{A}$ are initialized uniformly at random. The first row represents the $4$ first layers and the second row the $4$ last layers of Resnet-56.}
    \label{fig:HM56}
\end{figure*}
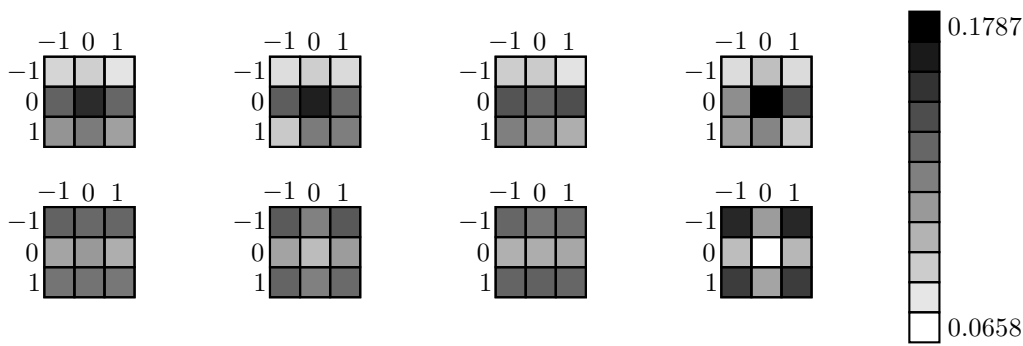

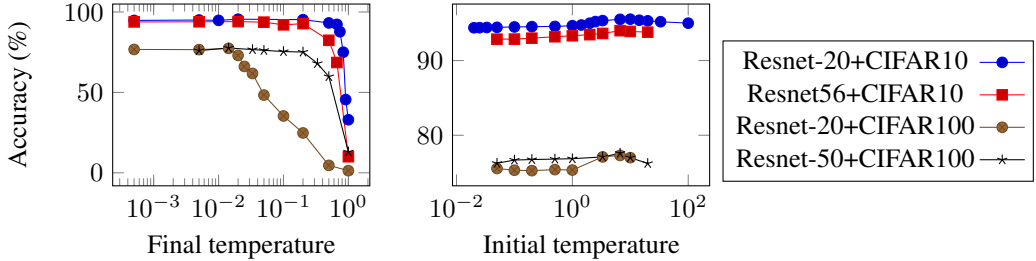
\begin{figure}[ht]
  \begin{center}
\begin{tabular}{cc}
    \begin{tikzpicture}\begin{axis}[xlabel=Final temperature,ylabel=Accuracy (\%),xmode=log, width=5cm,height=4cm,
legend style={at={(1.2,1)},anchor=north west,},
legend entries={},
legend plot pos=right]

\addplot coordinates {(1,33)(1/1.1,45.54)(1/1.2,75)(1/1.35,87.72)(1/1.5,92.3)(1/2,93.2)(1/5,95.23)(1/50,95.52)(1/100,94.79)(1/200,95)(1/2000,94.75)};
\addplot coordinates {(1,10)(1/1.5,68.6)(1/2,82.41)(1/5,92.75)(1/10,92)(1/20,93.7)(1/50,94)(1/200,93.9)(1/2000,93.8)};
\addplot coordinates {(1,1.34)(1/2,4.63)(1/5,24.79)(1/10,35.34)(1/20,48.34)(1/30,61.77)(1/40,66.14)(1/50,73)(1/70,77.39)(1/200,76.6)(1/2000,76.7)};
\addplot coordinates {(1,13)(1/2,59.82)(1/3,68)(1/5,75.08)(1/10,75.42)(1/20,76.15)(1/30,76.65)(1/70,77.6)(1/200,76)};
 \end{axis}
\end{tikzpicture}&
\begin{tikzpicture}
    \begin{axis}[xlabel=Initial temperature,xmode=log,width=5cm,height=4cm,
legend style={at={(1.05,0.8)},anchor=north west,},
legend entries={Resnet-20+CIFAR10, Resnet56+CIFAR10, Resnet-20+CIFAR100,Resnet-50+CIFAR100},
legend plot pos=right]
\addplot coordinates {(1/0.01,95.01)(1/0.03,95.2)(1/0.05, 95.32)(1/0.07,95.4)(1/.1,95.54)(1/.15,95.52)(1/.3,95.35)(1/.4,95.22)(1/.5,95)(1/.7,94.76)(1/1,94.67)(1/2,94.57)(1/5,94.54)(1/10,94.5)(1/20,94.45)(1/30,94.43)(1/40,94.4)(1/50,94.39)};

\addplot coordinates {(1/0.05,93.8)(1/.1,93.9)(1/0.15,94)(1/0.3,93.63)(1/0.5,93.47)(1,93.32)(1/2,93.19)(1/5,93)(1/10,92.88)(1/20,92.87)};
\addplot coordinates {(1/0.1,77)(1/.15,77.3)(1/.3,77.1)(1,75.34)(1/2,75.4)(1/5,75.26)(1/10,75.32)(1/20,75.55)};
\addplot coordinates {(1/0.05,76.23)(1/0.1,77)(1/0.15,77.6)(1/.3,77.1)(1,76.86)(1/2,76.81)(1/5, 76.75)(1/10,76.68)(1/20,76.25)};

\end{axis}
\end{tikzpicture}
\end{tabular}
  \end{center}
  \caption{Evolution of accuracy of Resnet-20/56 trained on CIFAR10 and Resnet-20/50 trained on CIFAR100 as function of final temperature (left), and as function of initial temperature (right).}
  \vspace{-0.45cm}
  \label{figure:final/initial}
\end{figure} 

In a third experiment, we observe the effect of initial and final temperature choices on accuracy. Figure~\ref{figure:final/initial}: left represents the evolution of accuracy of Resnet-20/56 trained on CIFAR10 and Resnet-20/50 trained on CIFAR100 as function of final temperature while initial temperature is fixed at $6.7$. It shows that the accuracy decreases when the final temperature becomes too high. Note that when the final temperature is large, obtained values in $\mathbf{A}$ at the end of the training process can be far from binary. In all cases, we round the values in $\mathbf{A}$ to the nearest integer before computing the accuracy. This experiment shows that final temperature values need to be small enough so the softmax can push the highest value to $1$ and the other values to $0$. Figure~\ref{figure:final/initial}: right shows the behaviour of the accuracy of Resnet-20/56 trained on CIFAR10 and Resnet-20/50 trained on CIFAR100 when initial temperature is changed and final temperature is fixed at $0.02$. We see an interesting region between $10$ and $6.7$ in which the accuracy is better.
It is worth mentioning that the choice of initial and final temperature is sensitive with respect to the obtained accuracy. Throughout our experiments, we observed that a too slow decrease in temperature causes the architecture to get stuck in local minima that are poorly fitted to the ending rounding operation. To the contrary, a too fast decrease in temperature prevents the learning procedure from finding the best shifts and boils down to an accuracy that is very similar to that of vanilla shift layers.


In the fourth experiment, we compare the accuracy, memory usage and FLOPs of SAL against vanilla Shiftnet and standard CNN on ImageNet ILSVRC 2012. Table~\ref{table:imagenet} shows that SAL is able to obtain better accuracies than vanilla Shiftnet and standard CNN for the same memory and FLOPs budget. 

  \begin{table*}
\centering

    \caption{Comparison of accuracy, number of parameters and FLOPs between a standard CNN, SAL and vanilla Shiftnet on ImageNet ILSVRC 2012.}
    \centering
    {\renewcommand{\arraystretch}{1.3}%
    \begin{tabular} { | c || c | c | c | c | c |} 

  \cline{3-6}
   \multicolumn{2}{l|}{} &  Top-1 & Top-5 & Params (M) & FLOPs (M) \\
  \hline
 Large &ResNet-w24 (CLs) & $63.47$ & $85.52$ & $\mathbf{3.2}$ & $664$   \\
 \cline{2-6}
budget &ShiftNet-A~\cite{wu2017shift} &$70.1$ & $89.7$ & $4.1$ & $1.4$G \\
\cline{2-6}
 &ResNet-w64 + SAL (ours) & $\mathbf{71}$ & $\mathbf{89.8}$ &  $3.3$ & $\mathbf{538}$ \\
   \hline
   \hline
 Small &  ResNet-w16 (CLs) & $56.6$ & $80.4$ & $1.4$ & $295$   \\
 \cline{2-6}
budget &ShiftNet-B~\cite{wu2017shift} &$61.2$ & $83.6$ & $1.1$ & $371$ \\
 \cline{2-6}
 &ResNet-w32 + SAL (ours) & $\mathbf{62.7}$ & $\mathbf{84}$ & $\mathbf{0.97}$ & $\mathbf{136}$ \\
   \hline

    \end{tabular}} \quad
    \vspace{.3cm}
  
  \label{table:imagenet}
  \vspace{-.5cm}
  \end{table*}

\input{final_figure.tex}

As last experiment, we propose to compare SAL with some other compression methods introduced in Section~\ref{sec:relatedwork}. To get a good approximation of memory needed to implement a neural network on a limited resources embedded system, we consider both weights and activations when one input data is processed, thus the memory footprint of a DNN will be the memory needed to store both weights and activations. 
 In our evaluation we compare SAL with Binary Connect (BC)~\cite{courbariaux2015binaryconnect} and Binary Weight Network (BWN)~\cite{rastegari2016xnor} applied to Resnet-20, MobileNetV2~\cite{sandler2018mobilenetv2} and Squeezenet~\cite{iandola2016squeezenet}. We also perform the comparison with another version of SAL denoted SAL2, in which we keep two weights per kernel instead of one. We use Resnet-20 with different number of weights and activations as baseline. Figure~\ref{fig:final_figure} shows that the baseline outperforms MobileNetV2, Squeezenet and applying BC on Resnet-20. It also shows that SAL and SAL2 outperform all other methods.

\section{Conclusion}
\label{sec:conclusion}
In this contribution, we proposed a novel attention-based pruning method that transforms a convolutional layer into a shift layer. The resulting network provides interesting improvements in terms of memory and computation time, while keeping high accuracy. Compared to existing methods, we showed that the proposed method results in improved accuracy for similar memory budgets as existing shift-layer based methods, using challenging vision datasets. The proposed method is thus an interesting alternative to channel-pruning methods.
 %


\bibliographystyle{unsrtnat}
\bibliography{biblio}
\end{document}

%% file: final_figure.tex
\begin{figure}[ht!]
  \begin{center}
    \begin{tikzpicture}
       \begin{scope}[scale=1]
        \begin{axis}[
            xlabel=Memory footprint,
            xmode=log,
            xmax=6100000,
            xticklabels={},
            extra x ticks={1000000,2000000,4000000},
            ymin=90,
            height=4cm,
            width=10cm,
            ylabel=Test set accuracy (\%),
            legend style={at={(1.03,1.08)},anchor=north west},
            legend entries={MobileNetV2, SqueezeNet, Resnet20, Resnet20+BC,Resnet20+BWN, Resnet20+SAL, Resnet20+SAL2},
            legend plot pos=right]
            
            ]
          \addplot coordinates
          {(5892452,94)
          };
          \addplot coordinates
          {(1469364,92.23)
          };
          \addplot coordinates
          {(4327754+1608970,95.21)(1084586+806026,94.4)(272474+404554,92.48)(68786+203818,88.37)
          };
          
          \addplot coordinates
          {(339394.375+2512282,94.34)(218817.5+2010442,93.5)(141328+1608970,93)
          };
          \addplot coordinates
          {(859225+4017802,95.7)(218817.5+2010442,95)(124590.625+1508602,94.27)(56713.75+1006762,93.41)(36941+806026,92.53)
          };
          \addplot coordinates
          {(15710+246322,83.2)(131410+732802,92.65)(231410+976042,93.64)(700010+1705762,95.01)(981609.5+2021974,95.3)(1421510+2435482,95.63)(2043010+2921962,95.8)
          };
          
          \addplot coordinates
          {(248860+732802,93.29)(685760+1219282,95)(985510+1462522,95.25)(2209960+2192242,95.86)(3297260+2678722,95.87)
          };

        \end{axis}
      \end{scope}
    \end{tikzpicture}
  \end{center}
  
  \vspace{-0.45cm}
  \caption{Evolution of accuracy when applying compression methods on different DNN architectures trained on CIFAR10.}
  
  \label{fig:final_figure}
\end{figure}
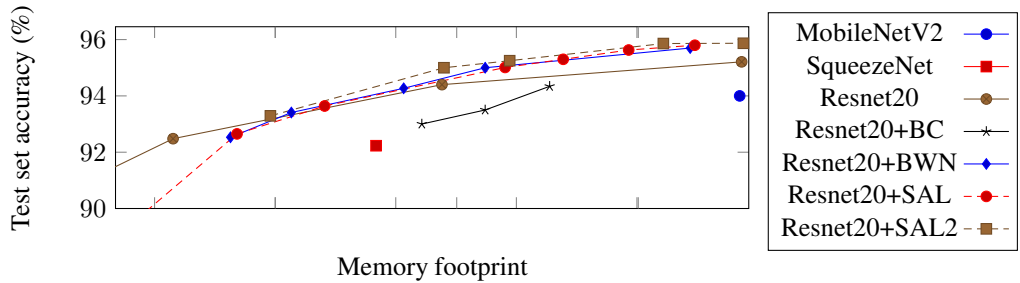